This pre-review manuscript of "Arabic Financial NLP" (Chapter 9) by Mo El-Haj has been accepted for publication in Arabic Digital Humanities, edited by Georgios Mikros, Mustafa Jarrar and Amal Haddad), John Benjamins Publishing Company, Natural Language Processing series (forthcoming).

# Arabic Financial NLP

**Mo El-Haj**
VinUniversity, Hanoi, Vietnam
Lancaster University, United Kingdom

## Abstract

Financial communication connects numerical accounts of economic activity with explanations of performance, responsibility and future expectations. Arabic financial natural language processing offers methods for studying these connections across corporate reports, market disclosures, journalism and banking interactions. Its contribution to digital humanities depends on retaining the relationship between a computational representation and the document from which it was made. This chapter follows that relationship from source collection and Arabic text processing to the analysis of vocabulary, evaluation, entities, causal attribution and multilingual reporting. It then examines what financial summaries and retrieval-grounded answers preserve or lose, and how dialectal intent detection changes the perspective from institutions speaking to institutions listening. Published Arabic resources are discussed in relation to the questions they were designed to answer, with a distinction between financial pretraining corpora, annotated datasets and evaluation benchmarks. A running constructed example connects the methodological discussion without presenting illustrative analysis as empirical findings. The chapter argues for an account of Arabic financial NLP in which linguistic context, documentary provenance and interpretive judgement remain part of the evidence throughout the analytical process.

## 1. Reading the words around the numbers

A company can report rising revenue and falling profit in the same announcement. Neither figure, considered alone, explains how the year will be remembered. The opening sentence may celebrate expansion, acknowledge pressure on profitability, or present the decline as a temporary consequence of investment. A subsequent news headline may retain only one of these elements. The numbers remain important, but their selection, arrangement and explanation shape the account that reaches a reader.

This relationship between numerical reporting and narrative is the starting point for Arabic financial natural language processing (NLP). Financial NLP applies computational methods to language associated with accounting, markets, banking and economic institutions. Financial narrative processing gives particular attention to the explanatory and evaluative language through which financial events are communicated. Arabic financial NLP examines these materials in Arabic and in multilingual settings where Arabic documents, terminology and interactions meet other languages. Its subject is a language domain, not a particular financial system: corporate earnings, banking services, market regulation and financial journalism belong within its scope regardless of the type of financial product being discussed. In their critical review of computational financial discourse research, El-Haj et al. (2019) identify a gap between the sophistication of financial questions and the linguistic simplicity of many text measurements. Their argument is useful here because it places computation within an interpretive process: counting expressions can locate a pattern, but understanding that pattern requires attention to word sense, context and the assumptions built into the measurement. They advocate complementarity between manual and computational analysis rather than a choice between them.

A digital humanities perspective develops this relationship further. Financial documents are records of institutional activity, but they are also authored representations of that activity. A report can reveal how an organisation names its stakeholders, distributes credit, explains setbacks and imagines its future. Its evidential value therefore extends beyond the extraction of facts into a spreadsheet. The question becomes how language participates in making economic events intelligible, credible or contestable, and how computational methods can help investigate that process.

Consider the following invented disclosure, used throughout this chapter. The company, figures and wording are constructed solely for analytical illustration; they are not taken from an actual report and do not constitute research findings.

> ارتفعت إيرادات الشركة من 100مليون ريال إلى 112مليون ريال، في حين تراجع الربح التشغيلي من 10ملايين ريال إلى 9 ملايين ريال .وأرجعت الإدارة هذا التراجع إلى ارتفاع تكاليف التشغيل، مع توقعها تحسناً في الأداء خلال العام المقبل، شريطة استقرار الطلب.

> The company’s revenue rose from 100 million riyals to 112 million riyals, while operating profit fell from 10 million riyals to 9 million riyals. Management attributed the decline in profit to higher operating costs. It expects an improvement next year if demand stabilises.

Four kinds of information coexist in this passage. There are reported quantities, a contrast between two performance measures, an explanation attributed to management, and a conditional expectation about the future. A system that assigns the whole passage a positive or negative label compresses those distinctions. A system that extracts only numbers loses the explanation. A summary that states that profit will recover replaces a conditional forecast with a certainty. These are different analytical failures, even when the resulting output is fluent.

The challenge is consequently more precise than making a computer “understand finance”. It is to decide which relationships a research question requires and how to preserve them as documents become searchable text, annotations, measurements and summaries. The account developed below follows this movement. It begins with the different texts through which an event circulates, considers how those texts become a corpus, and then examines the progressively richer representations used to study them. Banking interaction finally reverses the direction of communication, asking what happens when institutions interpret the language of the people who use their services.

## 2. One event, several financial texts

### 2.1 Genre changes what a statement does

The invented announcement could be reproduced in several forms. A stock-exchange disclosure might present the period’s figures and the stated reasons for change. A leadership statement might place the same results within a longer account of strategy. A financial journalist might foreground the profit decline and compare it with earlier expectations. A public comment might question whether revenue growth benefits shareholders or employees. These are plausible transformations of the example, not interchangeable observations of a single linguistic object.

Genre connects a text with a recurring communicative purpose, its producers and its anticipated audience. It matters computationally because the same expression can perform different work in different settings. A risk section records contingencies that may never occur; a news report can describe a realised loss; a customer message can request help with a pending transaction. Treating all three as instances of negative language would obscure the distinction between a possibility, an event and a request. Equally, the statement that a company "expects" improvement differs from an analyst's report that improvement "occurred". Attribution and time are part of the meaning, not optional metadata.

The work of El-Haj et al. (2020) shows why this matters at the level of documents. They develop and evaluate methods for retrieving, classifying and analysing narrative commentary in unstructured PDF annual reports. Their study concerns UK reporting, rather than an Arabic benchmark. Its methodological relevance is that a report must be divided into meaningful documentary components before a measure of narrative language can be interpreted. A chair's statement, an accounting policy and a financial-statement note should not acquire the same analytical status merely because they occupy adjacent pages.

For Arabic materials, this distinction supports more focused questions than a general survey of financial vocabulary. A study might compare explanations of declining operating profit in leadership statements and earnings announcements from the same institutions. Another might investigate how a central bank's account of uncertainty is reformulated in financial news. Corporate sustainability reporting invites questions about whether references to responsibility concern environmental commitments, employment, financial continuity or regulatory compliance. Such comparisons establish the relationship to be examined before choosing the method that will measure it.

### 2.2 Circulation, repetition and institutional voice

Documents also have relationships to other documents. An announcement can be quoted by a news agency, reproduced by several outlets and condensed into headlines. A report may repeat the wording of an earlier report. An official circular can amend a previous decision rather than stand alone. Source relationships are therefore necessary for distinguishing circulation from independent agreement.

Suppose the explanation in the constructed passage appears unchanged in ten news articles. Counting ten occurrences is appropriate for a question about its circulation. Counting ten independent explanations would be misleading. Deduplication is not simply the removal of unwanted repetition: the decision depends on whether repetition is noise for the research question or part of its subject. A corpus should retain links between originals, republications, quotations and revisions wherever these can be established. Institutional voice can then be studied as a relationship between what is said and who is represented as saying it. Expressions such as "management attributed", "the regulator stated" and "analysts expect" assign different sources to claims. Removing reporting clauses during extraction can make a journalist appear to endorse a company's explanation. Conversely, tracing those clauses can show how an institutional account travels with, or without, an explicit indication of its origin. This is a concrete route from information extraction to questions about authority and accountability.

### 2.3 Arabic and English as related documentary versions

Bilingual publication adds another relationship. Arabic and English files bearing the same corporate name and reporting year may be translations, adapted versions or partially overlapping publications. The relationship must be established from the documents; it cannot be inferred from filenames alone. Neither language should automatically be treated as the original, and a difference in wording is not sufficient evidence of deliberate audience targeting.

In the running example, an English rendering that preserves the figures but replaces "if demand stabilises" with "as demand stabilises" changes the status of the condition. One formulates a possibility; the other presents stabilisation as an accompanying circumstance. An embedding-based alignment system may regard the sentences as highly similar because almost all their content is shared. The difference that matters to a discourse analyst may nevertheless occupy only one short expression.

A useful comparison therefore proceeds from document identity to section correspondence, then to propositions and their wording. A proposition here means a claim that can be considered separately, such as revenue increasing or management expecting improvement. One Arabic sentence may correspond to two English sentences, so alignment should allow one-to-many and many-to-one relationships rather than force every sentence into a pair. Matching amounts, dates and institutional names can help locate corresponding passages, but these are anchors for inspection, not proof of equivalence.

The resulting analysis can distinguish a missing proposition from a change in modality, a terminological substitution or a shift in emphasis. These categories are more informative than a single score for bilingual similarity. They also make room for explanations that remain provisional: a difference may arise from translation, compression, document revision or editorial choice. Claims about the motives behind it require evidence beyond the textual difference itself.

## 3 From financial documents to analysable evidence

### 3.1 A corpus is a set of research decisions

A financial corpus is a deliberately assembled collection, not a synonym for every document that can be downloaded. Its boundaries determine the population about which an argument can be made. A collection of listed banks from one country can illuminate those institutions' reporting practices; it cannot by itself represent Arabic financial communication across sectors, countries and varieties.

Baker et al. (2008) provide a methodological precedent for connecting corpus patterns with contextual interpretation. Their study combines corpus linguistics and critical discourse analysis to examine UK press representations of refugees and asylum seekers. Although its subject is different, its relevance lies in the movement between quantitative patterns and close analysis of their textual contexts. Applied to financial discourse, this approach means using counts to guide investigation and using document reading to refine what should be counted.

For the running example, a defensible corpus might consist of matched earnings announcements and leadership statements, sampled within a stated sector and period. The unit of interpretation could be an explanation of performance, while the sampling unit remains the institution-year. Those units need not coincide. Several sentences from one report are still observations produced within the same institutional document, not several independent companies.

The corpus record should identify the institution, sector, reporting period, publication date, document genre, language, source and version. Reporting period and publication date are especially important to separate: a report issued in one year may concern the preceding year. Originals should be preserved, and passages should retain stable document identifiers and page or section locations. A checksum can establish that a file has not changed, but it cannot establish the accuracy of the institution's claims.

Missingness also belongs in the record. Earlier reports may be unavailable; one institution may publish only a shortened Arabic version; a scanned file may resist extraction. Excluding difficult files without recording the reason can turn technical accessibility into an unacknowledged sampling criterion. Comparing the distribution of included and missing documents is therefore part of understanding the corpus, rather than an administrative task completed before analysis begins.

### 3.2 Layout is part of the evidence

The step from PDF to text is a form of transcription. A financial page can combine paragraphs, two-column layouts, headings, tables, footnotes and captions. Meaning depends on their relationships. A number printed as 112 beneath a heading that specifies "millions of riyals" has a different interpretation if the heading disappears. A negative sign lost during conversion can reverse a conclusion even when almost every other character is correct.

Optical character recognition (OCR) is required for text stored as an image, whereas a PDF with a usable text layer may permit direct extraction. Both routes require checking. An embedded text layer can still produce the wrong reading order, duplicate headers or detach table labels. The relevant test is correspondence with the page, including how Arabic and Latin fragments are ordered, rather than the mere presence of selectable characters.

Kang et al. (2022) formulate document-structure extraction as the detection of headings and their organisation into a table of contents. FinTOC 2022 includes English, French and Spanish financial documents, and its successful submissions combine textual, structural and visual features. This is evidence for the usefulness of layout-aware processing, not a reported evaluation on Arabic. An Arabic corpus needs its own checks for right-to-left order, mixed scripts, headings and numerical conventions. An appropriate representation retains several layers: the original page, its transcription, and an analysis version linked back to the transcription. Tables should retain row labels, column labels, units and notes rather than become isolated strings of numbers. A sample-based extraction audit should include difficult pages as well as clean running prose. Otherwise, a low average character error rate can disguise systematic mistakes in the very figures on which a financial interpretation depends.

### 3.3 Arabic words are not fixed computational units

Arabic processing introduces further choices within the text itself. Habash (2010) explains why morphology, orthography and the relationship between Arabic varieties are central to computational analysis. Conjunctions, prepositions and pronominal elements may attach to other forms, while the absence of short vowels leaves some distinctions to context. A visible word, a morphological unit and a model's subword token are therefore different objects.

For example, the constructed form وبمساهمتهم, "and with their contribution", contains elements that can be analysed separately. A word-form count keeps the written form intact. Clitic segmentation separates attached grammatical elements. Lemmatisation groups inflected forms under a dictionary form. Root-based stemming can group a still wider family, but that family need not correspond to one financial concept. The best unit depends on the question: a study of pronouns and agency requires different distinctions from a search for variant forms of a technical term.

Obeid et al. (2020) describe CAMeL Tools, a toolkit that provides Arabic preprocessing, morphological modelling and other language-analysis components. Its relevance is the availability of reusable linguistic analysis, rather than a requirement to adopt every component of a standard pipeline. Automated disambiguation should be inspected on the financial material being studied. A linguistically plausible analysis can still be wrong in a specialised expression or an institutional name.

Normalisation should likewise be selective. Unifying selected alef variants can improve retrieval, but merging forms indiscriminately can erase distinctions in names or terminology. Removing all punctuation can damage decimal values and percentages. Removing Latin characters can delete currency codes, instrument identifiers or acronyms. Generic stopword removal can discard لم and لن, or expressions of expectation and attribution, even though negation and modality are essential to the running example. The original form should remain recoverable from the normalised one.

Arabic financial terminology further resists simple substitution. الالتزامات المالية can denote financial liabilities or obligations, whereas الالتزام بالمعايير concerns compliance with standards. The shared lexical material does not make these equivalent categories. A glossary becomes analytically useful when it records expressions, contexts and institutional usage, rather than imposing one English equivalent on every occurrence. Terminological variation can then be studied without being mistaken for processing error.

### 3.4 A financial fact contains more than a value

Table 1 makes explicit the quantities in the invented announcement. The earlier and later periods are deliberately generic, and no currency jurisdiction is assigned beyond the wording "riyals".

**Table 1:** Numerical content of the constructed announcement. Values are in millions of riyals. Changes are calculated from the invented figures, not empirical observations.

| Metric | Earlier period | Later period | Relative change |
|---|---|---|---|
| Revenue | 100 | 112 | +12% |
| Operating profit | 10 | 9 | −10% |

The revenue change is $(112-100)/100 = 0.12$. The operating-profit change is $(9-10)/10 = -0.10$. These operations are straightforward only because the quantities have already been associated with the correct measure, scale and comparison period. A value of 9 without those associations is not a usable statement about profit. Nor does a decline in profit mean that the company made a loss: in this example, operating profit remains positive.

A structured record should therefore connect the value with its metric, unit, period, entity and source location. It should also distinguish a value directly reported in the text from one calculated by the researcher. In the example, the percentage changes are derived from the stated amounts. Recording the derivation makes the result inspectable and prevents a calculated quantity from later being presented as an institutional quotation.

Chen et al. (2021) make this connection central to FinQA, an English-language dataset of expert-written questions over financial reports accompanied by reasoning programmes. The task requires relevant evidence and numerical operations, rather than text matching alone. It provides a model for specifying evidence-linked financial reasoning, but no direct evidence of Arabic performance. For Arabic research, the corresponding challenge includes the preservation of Arabic-Indic and Western digits, mixed-script labels and local presentation conventions before any operation is attempted.

The interpretive consequences extend beyond arithmetic. A relative change is undefined when its comparison denominator is zero, and negative comparison values require particular care in description. Differences between percentages may instead need to be expressed in percentage points. These distinctions show why financial meaning cannot be recovered by extracting digits and leaving their relationships for later.

## 3.5 From words to patterns

### 3.5.1 Concordances connect a pattern to its contexts

Once source relationships and textual units have been preserved, the first useful analysis may be a simple search. A concordance displays occurrences of an expression with their surrounding words and links to the source passage. It can reveal whether ضغوط (pressures) concerns costs, liquidity, supply or public criticism, and whether تحسن (improvement) describes an observed result, an expectation or a denied development. These distinctions often become apparent before a more elaborate model is needed.

The corpus-linguistic account of McEnery and Hardie (2012) distinguishes frequency, concordance, keyword and collocation analysis as related ways of examining patterned language use. Their value in the present setting is the ability to move between a distribution and the occurrences that constitute it. A frequency list suggests what recurs; a concordance reveals how it is used; a comparison identifies where the recurrence is distinctive.

A raw count needs a denominator. If a term occurs 60 times in a collection of 30,000 tokens and 80 times in a collection of 80,000 tokens, its relative frequencies are 20 and 10 per 10,000 tokens respectively. These invented counts illustrate that the larger absolute count need not indicate greater concentration. The comparison also depends on using the same tokenisation convention. Arabic segmentation that increases the number of tokens can alter a normalised rate even when the underlying passages are unchanged.

Frequency alone does not show dispersion. Sixty occurrences concentrated in one institution's repeatedly reproduced statement differ from sixty occurrences spread across many institutions. A diachronic study should therefore inspect both the rate of use and the number of documents or institutions in which the expression appears. This helps distinguish a widespread change in vocabulary from the expansion of one prolific source.

Keyword analysis asks which items occur disproportionately in a target collection relative to a reference collection. "Key" is a statistical description of that comparison, not a guarantee of conceptual importance. Comparing leadership statements with accounting notes may mainly reveal genre differences. Comparing equivalent sections across matched periods is more suitable for a question about historical change. Effect sizes, frequencies and source dispersion help interpret statistical significance, particularly when large collections make small differences appear highly significant.

Collocation analysis extends the question from individual expressions to associations. It can investigate whether النمو (growth) occurs near words for revenue, employment or national development, or which terms appear around المخاطر (risks). The association depends on the context window, segmentation and statistic used. A narrow window emphasises local phrase patterns; a wider one may capture broader thematic association. In either case, the result identifies textual proximity rather than a causal relationship.

For the constructed announcement, concordance searches for عزت, بسبب and نتيجة could locate candidate explanations. Those expressions do not exhaust Arabic causal language, and their occurrence does not always introduce a causal claim. Their practical role is to produce a retrievable set of passages whose linguistic functions can be examined. Search expansion through observed variants is then guided by evidence rather than by a fixed dictionary assumed to be complete.

### 3.5.2 Topics organise an archive, not an argument

When a collection is too varied to navigate through a few known terms, topic modelling can provide an exploratory map. In Latent Dirichlet Allocation (LDA), introduced by Blei et al. (2003), each document is represented as a mixture of topics, and each topic as a probability distribution over words. A model might group expressions associated with lending and liquidity separately from expressions associated with employment and training. The analyst supplies an interpretation of the word distribution; the model does not discover a pre-existing, self-explanatory heading.

The unit of analysis matters here. An entire annual report may combine many unrelated sections, whereas a paragraph may be too short to provide stable evidence of co-occurrence. Section-level modelling can be a useful compromise when section boundaries are reliable. A topic dominated by audit terminology may describe document structure rather than a shift in institutional priorities. Repeated templates can similarly produce apparently coherent topics that say little about substantive change.

Interpreting a topic requires reading both highly associated passages and less typical ones. Changing the number of topics, preprocessing choices or random seed can reveal whether an interpretation depends on a fragile modelling decision. Comparing topics across periods requires a shared or explicitly aligned representation; independently fitted "topic 3" in two models does not necessarily denote the same theme. A plotted trend is meaningful only after this correspondence has been established.

Embedding-based clustering provides a different exploratory route. Instead of modelling word counts in LDA's probabilistic form, it can group passages according to similarities between their vector representations. It should not be described as if it were the same statistical model. Its usefulness likewise depends on whether the resulting groups distinguish the phenomena of interest, rather than simply separating publishers, formats or formulaic language.

### 3.5.3 Representations bring the domain into the model

An embedding is a numerical representation that makes similarity between words or passages computable. A static word embedding assigns one vector to a word form, combining evidence from its uses. Contextual representations instead depend on the surrounding passage, allowing the representation of a term to vary with its use. Neither representation is a definition of the term; both are learned from patterns in training data.

Zmandar et al. (2021) developed financial word embeddings for Arabic, English and French. Their contribution was to provide representations learned from financial language rather than relying exclusively on general-domain vocabulary. The reported downstream sentiment experiment used English financial sentences, so it should not be reported as a demonstrated Arabic classification improvement. Nor does the provision of embeddings in several languages by itself establish that their vector spaces are aligned for direct cross-language comparison.

For a terminology study, such representations can suggest related expressions that an initial glossary missed. Similarity still requires inspection: "profit" and "loss" may share contexts without being synonyms. A term's nearest neighbours can reflect corporate boilerplate, frequency or publisher style as well as financial meaning. Changes in neighbours across time can therefore suggest an interpretive question, but do not by themselves demonstrate a change in the underlying concept.

Transformer-based encoders provide contextual representations for tasks such as classification and entity recognition. Antoun et al. (2020) introduced AraBERT for Arabic language understanding; Abdul-Mageed, Elmadany et al. (2021) developed ARBERT and MARBERT with different coverage of formal and social-media Arabic. These resources make it possible to begin from a learned Arabic representation instead of constructing every feature manually. They are general Arabic models, however, rather than evidence that every specialised financial expression or dialectal query will be interpreted correctly.

The distinction between pretraining and task learning is important. Pretraining develops representations from large text collections. Fine-tuning adjusts a model to labelled examples or a particular task. Continued domain pretraining exposes an existing model to financial text, while task-specific fine-tuning teaches a particular decision or output format. These steps answer different needs. A model may become more familiar with financial wording without acquiring a reliable rule for assigning attribution labels or preserving numerical facts.

## 3.6 From patterns to claims about performance and responsibility

### 3.6.1 Financial evaluation is not a single sentiment

The next question is what the patterns mean. Sentiment analysis commonly assigns polarity labels such as positive, negative and neutral. Financial discourse requires the object and perspective of evaluation to be specified. An increase in borrowing costs can be unfavourable for one actor and favourable for another. A fall in expenses and a fall in revenue share a direction of change but not the same financial interpretation. A calm sentence can report an adverse result without expressing an emotion.

Loughran and McDonald (2011) demonstrated the problems created by general-purpose word lists in English financial filings and developed domain-specific textual categories. Their study concerns US 10-K reports, not Arabic language validation. Its contribution here is a principle of measurement: apparently negative vocabulary can have a technical financial function. Translating an English dictionary into Arabic would not resolve polysemy, morphology, negation or differences in genre. The resulting resource would still require validation on Arabic passages.

Several distinctions can be maintained without making the analysis unnecessarily complicated. *Performance direction* concerns the change being reported. *Financial evaluation* concerns whether that change is favourable for a stated target or perspective. *Affect* concerns expressed feeling. *Stance* concerns a position towards a proposition, policy or actor. *Modality* concerns possibility, expectation, necessity or commitment. An annotation scheme need not label all of them, but should not quietly use one as a substitute for another.

Two shortened versions of the constructed example illustrate why discourse structure matters:

رغم انخفاض الربح التشغيلي، ارتفعت الإيرادات.

Despite the decline in operating profit, revenue rose.

رغم ارتفاع الإيرادات، انخفض الربح التشغيلي.

Despite the increase in revenue, operating profit fell.

The propositions are shared, but the concessive construction changes their arrangement. In the first version, declining profit provides the background against which revenue growth is asserted. In the second, revenue growth provides the background against which the decline is asserted. This is an analysis of the constructed wording, not a measured claim about its psychological effect on readers. Establishing such an effect would require separate evidence.

Social-media polarity is a related but distinct object. BORSAH, introduced by Alshahrani et al. (2018), supplies Arabic financial tweets annotated for subjectivity and sentiment. It establishes financial social media as a resource for Arabic opinion analysis. Its labels and platform setting should not be transferred unquestioningly to corporate explanations of performance. A public post, an earnings announcement and a leadership statement can all concern the same event while requiring different annotation decisions.

### 3.6.2 Attribution records an explanation, not a proven cause

The statement that management attributed declining profit to higher costs makes a claim about causation and identifies its source. An extraction system should preserve both. If it returns only “higher costs caused declining profit”, it has removed the attribution and made the explanation appear independently established. For a discourse study, this is a substantive change to the evidence.

El-Haj et al. (2016) operationalise this distinction through annotated UK preliminary earnings announcements, examining performance, tone and internal or external attribution. Their study investigates whether computational models can reproduce expert coding of those categories at scale. It provides a concrete model of turning a question about corporate explanations into annotation and classification. It does not provide a ready-made Arabic attribution scheme, nor establish that an external explanation is necessarily misleading.

In Arabic material, candidate constructions include explicit reporting verbs, causal expressions and descriptions of agency. Each requires contextual interpretation. “Market conditions” may be presented as external, but the mere presence of a market term does not determine the explanation’s classification. A sentence may quote a rejected explanation or contrast several possible causes. An apparent absence of agency may reflect genre convention or sentence structure rather than an attempt to conceal responsibility.

Table 2 separates the propositions in the running example. The categories are an illustrative analytical design, not labels released with an existing dataset.

**Table 2:** An illustrative proposition-level representation of the constructed disclosure. Evidence status is retained separately from the subject matter of the claim.

| Proposition | What can be represented | What should not be inferred |
| --- | --- | --- |
| Revenue rose from 100 to 112 million riyals. | Metric, two values, periods, unit and reported increase. | That every measure of performance improved. |
| Operating profit fell from 10 to 9 million riyals. | A decline in a specified positive profit measure. | That the company recorded an operating loss. |
| Management attributed the decline to operating costs. | Claim source, proposed cause and stated effect. | That the causal account has been independently verified. |
| Management expects improvement if demand stabilises. | Forecast source, future orientation and explicit condition. | That improvement or demand stabilisation is certain. |

A supervised classifier learns to apply such categories from annotated examples. A simple baseline might combine word and character features with logistic regression; a contextual model uses learned passage representations. The comparison is useful only when the categories and evaluation set are fixed. Character features can be helpful for some spelling variation, but can also learn source-specific patterns that do not generalise. A more complex model should therefore be judged by the errors it resolves, not by its architecture alone.

Annotation is part of this reasoning. Guidelines should distinguish asserted, negated, hypothetical and quoted claims, specify how targets and speakers are identified, and allow genuinely unclear cases. Agreement among annotators tests whether a scheme can be applied consistently; it does not establish that the scheme adequately represents responsibility or optimism. Disagreement can reveal a theoretical distinction that the initial label inventory omitted.

### 3.6.3 Entities and events connect claims across texts

To compare explanations across documents, the actors and measures must also be identified. Named entity recognition (NER) locates spans such as organisations, people, currencies and dates. Financial applications may need distinctions between companies, regulators, instruments, indicators and other specialised categories. Entity linking then associates a mention with a particular referent, so that abbreviated names or bilingual variants do not automatically become separate institutions.

AMWAL, presented by Abdo et al. (2025), addresses Arabic financial-news NER using material from three news sources covering 2000–2023. The paper reports 17,185 annotated financial entity tokens across twenty categories, with a scheme informed by the Financial Industry Business Ontology and supplemented for the task. Its semi-automatic annotation and manual review illustrate how a specialised domain requires decisions beyond a generic person–organisation–location inventory. The underlying collected news corpus and the annotated entity material are different quantities; the size of one should not be reported as the size of the other.

These distinctions support questions about whose activity becomes visible in financial discourse. Entity mentions can identify which organisations receive attention and which actors are repeatedly quoted. They do not establish ownership, influence or cooperation. Two institutions appearing in the same article may be competitors, participants in a transaction or merely examples in a market overview. A co-occurrence network records textual association unless a relation has been explicitly extracted and validated.

Relation extraction identifies a stated connection, such as one organisation acquiring another. Event extraction adds an occurrence and its participants, time and other arguments. Linking events across an announcement and a news report can support analysis of reframing, but shared names and dates are insufficient when several events concern the same firm. Quantities, event type and reporting period provide additional constraints. The resulting record should retain which source reported each relation rather than collapse all accounts into one apparently uncontested history.

At this point, computation has moved beyond counting words to representing claims. That richer representation makes summarisation and question answering possible, while also clarifying what those systems must not discard.

## 3.7 What summaries and generated answers retain

### 3.7.1 Compression makes a choice about importance

A summary is a selective representation of a source. An extractive system chooses existing sentences, whereas an abstractive system generates new wording. Extractive selection preserves the wording it retains, but can omit a qualification or separate a pronoun from its antecedent. Abstractive generation can combine dispersed information more naturally, but can also change quantities, relationships or the status of a claim. Neither approach is faithful merely because it reads well.

The purpose of the summary matters. A headline, an overview of a long report and a digest of changes since the previous year have different selection criteria. Evaluating them against the same idea of "the important information" would obscure these differences. A headline may foreground one development; a research summary comparing performance and attribution needs to preserve both the adverse result and the explanation offered for it.

Nagoudi et al. (2021) introduce AraT5 as a family of Arabic text-to-text transformer models evaluated on language-generation tasks. Text-to-text modelling expresses tasks through an input text and an output text, making it suitable for generation as well as other transformations. This provides the general Arabic modelling basis for subsequent domain adaptation.

Zmandar et al. (2023) develop FinAraT5 through continued pretraining of AraT5 on Arabic financial text, including annual and quarterly reports and press releases. Their associated FinAraSum dataset contains 48,900 financial and economic news article–title pairs from Al Arabiya Aswaq, covering 2012–2021, divided into 44,900 training, 1,500 validation and 2,500 test examples. The targets are short, headline-style summaries. The study therefore provides an Arabic financial model and a news-generation evaluation resource, but its title-generation results should not be described as validation of faithful long-report summarisation. The financial pretraining corpus and the labelled summarisation dataset serve different purposes and must remain distinct in accounts of the work.

The distinction matters when a resource is reused for humanities research. A model trained to reproduce headlines learns a particular form of selection. Applying it to an annual report does not establish that it preserves the report's balance between growth, adverse performance and uncertainty. The target genre changes what omission means. A useful adaptation begins by specifying the intended summary, its source evidence and the kinds of distinction that must survive compression.

For the running disclosure, an acceptable constructed summary would be: "Revenue rose, but operating profit fell; management linked the decline to higher operating costs and made its expectation of improvement conditional on stable demand." This omits the amounts but retains the contrast, source of explanation and condition. Where numerical comparison is the purpose, the amounts or correctly calculated changes should also be included. By contrast, "The company improved performance and expects profit to recover" broadens a favourable claim and removes the adverse result and explicit condition.

### 3.7.2 Similarity is not the same as fidelity

Lin (2004) introduced ROUGE as a family of automatic measures comparing a system summary with reference summaries through overlapping textual units. It is useful for assessing aspects of content overlap at scale. Its dependence on tokenisation and matching means that Arabic segmentation and orthographic conventions affect the comparison. It also means that preserving many words does not guarantee preservation of their relations.

Zhang et al. (2019) propose BERTScore, which compares generated and reference texts using contextual token representations. This can recognise correspondences that exact word matching misses. It is still a similarity measure rather than a direct verification of every financial assertion. A changed number, negation or condition can be consequential even when the surrounding wording remains highly similar.

Financial evaluation should consequently examine several separable properties: coverage of the required propositions, support in the source, numerical consistency, attribution and preservation of uncertainty. The constructed summary can be checked proposition by proposition against Table 9.2. A correct amount paired with the wrong metric is an error. A plausible explanation absent from the source is unsupported. An explanation correctly attributed in the source but asserted without qualification in the summary has changed evidential status.

This form of checking also distinguishes factual truth from source faithfulness. A generated statement might be true according to another document but unsupported by the document it summarises. Conversely, a faithful summary can accurately report an institution's claim without establishing that the claim is true in the world. Humanities analysis needs this distinction because its object may be the historical production of a claim, including a claim that was later challenged.

### 3.7.3 Retrieval makes sources available, not infallible

Question answering over a financial archive adds a retrieval problem. Before an answer can be generated, the system must locate relevant evidence. A query about the profit decline may require the results table, a management explanation and a footnote defining the comparison. Retrieving only the sentence containing "profit" is not necessarily sufficient.

Lewis et al. (2020) introduced a retrieval-augmented generation approach that combines a pretrained generator with retrieved passages from an external collection. Their experiments use knowledge-intensive NLP tasks, rather than Arabic corporate archives. The approach motivates a useful separation: information can be retrieved from identifiable documents instead of being supplied solely from a model's parameters. In a financial research setting, that collection can be restricted to documented sources, versions and periods.

Document division then becomes important again. A passage prepared for retrieval should retain its heading, source identifier and relevant table context. A fixed-length chunk ending before "if demand stabilises" can make the conditional forecast unavailable to the generator. A retrieved current-year table paired with a previous-year explanation can produce a coherent but mismatched answer. Temporal and documentary identity must therefore accompany textual relevance.

Gao et al. (2023) develop ALCE to evaluate language-model answers with citations, considering the quality of the answer and its evidential support. Their work makes a distinction directly relevant to financial archives: a citation's presence is not equivalent to its correctness or its completeness. An answer may cite a real page that does not support the attached claim, or support one clause while leaving another unsupported.

Retrieval and generation should accordingly be tested separately. Retrieval recall asks whether required evidence is found within the returned set. Answer evaluation asks whether the generated propositions follow from that evidence. Citation evaluation asks whether the references support the relevant propositions and cover the claims that require support. Unsupported questions should remain unanswered or explicitly unresolved. A system that reliably returns the right passage can already be useful for scholarship without being authorised to produce a definitive interpretation.

## 3.8 When financial institutions listen

### 3.8.1 From institutional narrative to customer intention

The preceding sections concern institutions describing their activity. Banking interaction changes the direction of analysis: an institution must interpret a person's request. The connection is deeper than the shared financial vocabulary. In both settings, a computational representation determines which distinctions remain visible. A summariser decides what to retain from a report; an intent classifier decides which institutional category best represents a customer's message.

Intent detection assigns a query to a defined purpose, such as a problem with a card, a transfer or a charge. Consider two constructed expressions of a payment-card problem:

لم تصل بطاقتي بعد. بطاقتي لسه
ما وصلت.

My card has not arrived yet.

The first uses a more formal written construction; the second uses a colloquial formulation found in Levantine usage, without being a diagnostic marker of a single community. Both can express a card-delivery problem. The variation matters because an institution should not require one written register as the price of having a request recognised. Yet an isolated query may still be ambiguous about whether the card is new, renewed or a replacement, and a confident classification cannot supply missing context.

Jarrar et al. (2023) address this area with ArBanking77, an Arabic adaptation of the English Banking77 intent inventory. The original published resource contains 31,404 queries in Modern Standard Arabic and Palestinian Arabic across 77 intents, using localisation and expansion rather than treating literal translation as sufficient. It enables controlled investigation of intent detection under linguistic variation. Its construction should not be confused with a representative sample of naturally occurring conversations from Arabic-speaking bank customers.

Malaysha et al. (2024) extend the setting through AraFinNLP 2024, with tasks for multidialect intent detection and cross-dialect translation with intent preservation. Their published split table totals 39,315 query instances across Modern Standard Arabic, Palestinian, Moroccan, Saudi and Tunisian Arabic. This is not a collection of 39,315 distinct meanings, each represented in every variety. The training partition contains Modern Standard Arabic and Palestinian queries, while the other varieties appear in the test partition, with unequal coverage. The distinction is important because the task involves cross-variety transfer rather than equally resourced training in five varieties.

Together, these resources reveal why a benchmark's unit, partition and provenance matter. Several dialectal versions can represent the same underlying query. Their labels establish a controlled comparison but do not tell us how often that customer problem occurs in everyday banking. Class frequencies in a designed benchmark cannot be used as estimates of social prevalence without independent evidence.

### 3.8.2 A correct category may still be an incomplete account

An intent inventory is an institutional taxonomy: it defines the kinds of request a system is designed to recognise. A message can contain several requests, provide an explanation before asking for help, or describe a difficulty outside that inventory. Forcing every message into one of the known classes can make an unsupported interpretation appear successful simply because a label was produced.

This is where classification meets questions of access and representation. A benchmark can test whether two varieties are assigned equivalent task labels. It does not establish that the system understands their tone, politeness, urgency or social meaning. Likewise, preserving intent in translation does not entail preserving every feature of the interaction. What counts as equivalence must be defined in relation to the claim being made.

Inoue et al. (2021) examine the relationship between Arabic language-model variant, pretraining-data size and task type. Their comparisons show why model selection should consider the language variety and task, rather than assume that a larger pretraining corpus is universally preferable. For financial interaction, this implies testing on the kinds of queries the institution expects to receive, including indirect requests, spelling variation and code-switching where relevant.

Accuracy reports the proportion of correct classifications, but can conceal poorly recognised categories. Precision for an intent measures how often predictions of that intent are correct; recall measures how many true instances are found. Their harmonic mean is F1, and macro F1 averages the class-level F1 scores with equal weight. Reporting these measures by variety and intent helps identify concentrated errors, provided the sample sizes and composition are also reported. Differences between unmatched groups should not automatically be attributed to dialect alone.

Error analysis then returns to language. Confusion between a delayed transfer and a failed transfer may arise from tense, aspect, negation, vocabulary or genuinely missing information. An evaluation should also consider requests outside the label inventory and the option to seek clarification rather than force a category. A model's numerical confidence is not automatically a calibrated probability of being correct. The relevant institutional question is whether the system can recognise its limits without placing disproportionate effort on users whose language is less well represented.

## 3.9 From a computational result to a defensible interpretation

The methods now form a connected investigation rather than a collection of independent tools. A source is preserved and situated within its genre; a passage is represented without losing its quantities and claims; patterns are located across comparable documents; and selected findings are returned to their contexts. The final step is to establish what argument this process supports.

A bounded study of the running theme could examine how companies explain declining operating profit in Arabic announcements and leadership statements, and how the explanation changes when reproduced in news or another language. This is a proposed research design, not a report of an experiment conducted for this chapter. The comparison would begin with matched institution-period records and documented source relationships. Numerical verification would identify the relevant performance change. Concordances would locate candidate explanations, and annotation would distinguish their sources, causes, modality and rhetorical arrangement.

A model could then extend the annotation across a larger corpus, but its evaluation must reflect the intended inference. Randomly assigning neighbouring sentences from the same report to training and test sets risks measuring familiarity with a document rather than generalisation. Repeated templates, translated versions and republications can create similar leakage. Holding out institutions, later periods or document families tests different questions and should be chosen accordingly. Future-period evaluation also requires excluding later information from the evidence available at prediction time.

Errors must be connected to their consequences for interpretation. If a system disproportionately misses indirect explanations of poor performance, aggregate counts may exaggerate the prominence of explicit external attribution. If Arabic PDFs lose more qualifying clauses than English PDFs, an apparent translation difference may be an extraction artefact. A reliable argument therefore reports uncertainty in the observations as well as variation in the phenomenon under study.

Robustness checks can ask whether a pattern persists when repeated templates are retained or separated, when a different reference corpus is used, or when uncertain annotations are excluded. Because many observations come from the same institution or document, uncertainty estimates should respect that grouping rather than treat every sentence as independent. Close reading should include typical passages, exceptions and cases that challenge the initial explanation. Selecting only examples that confirm a model's dominant pattern would make interpretation circular.

The resulting claim may be narrower than a striking headline suggests. Finding more external explanations in a set of adverse-performance announcements supports an account of the language used in those documents. It does not demonstrate deception, establish management's private motives or prove the economic causes of performance. Establishing the latter would require other evidence and a different research design. The strength of the humanities contribution lies in explaining the pattern's institutional and linguistic significance without exceeding its evidential basis.

Documentation makes that basis available for scrutiny. Bender and Friedman (2018) propose data statements to describe the populations and linguistic circumstances represented in NLP datasets, aiming to improve the precision of generalisation and address exclusion. For Arabic financial research, a corresponding statement should identify countries, genres, varieties, periods, source-selection criteria and known absences. Annotation guidance, processing decisions, model versions and prompts should explain how the source was transformed. Where documents cannot be redistributed, permitted metadata and derived records can still help others understand the analysis without implying unrestricted reuse rights.

The distinction between public institutional documents and personal financial communication must also remain visible. Customer messages can contain account details, contact information or descriptions of financial difficulty. Removing names does not necessarily remove the possibility of identification through combined details. Collection, access and publication should therefore minimise exposure and follow applicable institutional and data-governance requirements. Synthetic examples can illustrate an analytical distinction safely, as they do here, but cannot substitute for empirical evidence about the distribution or difficulty of real language.

The most productive future resources would connect capabilities that existing datasets tend to separate: page-grounded Arabic text, numerical records, attributed explanations, bilingual document relationships and source-linked summaries. This is a research direction motivated by the distinctions above, rather than a claim that no related work exists. Such resources would make it possible to evaluate whether a system preserves meaning across a sequence of transformations, instead of asking only whether it performs one isolated task well.

### 3.10 Conclusion

The invented announcement began with two numbers moving in different directions. Following it through disclosure, interpretation, translation and compression has shown why Arabic financial NLP needs more than a polarity label or a plausible summary. Revenue growth, declining operating profit, an attributed cause and a conditional expectation are different kinds of evidence. Their relationships must survive the transformations that make large-scale analysis possible.

Arabic financial NLP contributes to digital humanities by making those transformations inspectable. Corpus methods reveal distributions while preserving access to individual occurrences. Linguistic annotation distinguishes targets, speakers and the status of claims. Entity and event analysis connects passages without turning textual association into an assumed financial relationship. Summarisation and retrieval improve access to an archive when their outputs remain accountable to it. Dialectal banking analysis extends the same concern to the institutional categories through which customers are heard.

The central scholarly movement is therefore from documents to representations and back again. A computational result becomes informative when its categories are justified, its source can be recovered and its interpretation respects the limits of the evidence. The aim is not to replace the financial text with a score, but to understand more precisely how Arabic language gives economic activity a public meaning.